\documentclass{article}

\usepackage[dblblindworkshop, final]{neurips_2026}
\workshoptitle{First Workshop on Reliable Evaluation for Language Models (JUDGe)}

\usepackage[utf8]{inputenc}
\usepackage[T1]{fontenc}
\usepackage{hyperref}
\usepackage{url}
\usepackage{booktabs}
\usepackage{amsfonts}
\usepackage{amssymb}  
\usepackage{amsmath}
\usepackage{nicefrac}
\usepackage{microtype}
\usepackage{xcolor}
\usepackage{graphicx}

\title{Agreement With Humans Is Not a Property of the Judge}

\author{%
  Diya Narayanan \\
  Birla Institute of Technology and Science, Pilani \\
  \texttt{diya.narayanan22@gmail.com} \\
}

\begin{document}

\maketitle

\begin{abstract}
LLM judges are commonly evaluated by correlating their scores with human ratings. This correlation is often interpreted as evidence of how well the judge evaluates outputs. Using the two annotator panels SummEval
\citep{fabbri2021summevalreevaluatingsummarizationevaluation} released over the same 1{,}600
summaries, one judge's scores correlate 0.43--0.61 with the three experts and only 0.03--0.07 with
the five crowdworkers. The crowd panel is not noise: a three-expert panel predicts it no better than
the judge does, and scaling the judge $64\times$ does not close the gap. Both panels' aggregates are
reliable, so correcting for annotator noise declares both targets sound and reliability is blind to
the failure. Since judge scores increasingly gate model comparisons, an agreement figure reported
without naming the panel it was computed against cannot carry the conclusion it is usually asked to
support.
\end{abstract}

\section{Introduction}

\looseness=-1
LLM judges are evaluated by correlating their scores with human ratings, often treating agreement as a property of the judge. Yet benchmarks typically merge annotator ratings into one gold score, obscuring which humans the judge agrees with.

\looseness=-1
SummEval, a summarisation benchmark that preserves separate ratings from expert and crowdworker panels on the same summaries provides an opportunity to investigate this question. Its authors found almost no correlation between the two panels’ aggregate ratings \citep{fabbri2021summevalreevaluatingsummarizationevaluation}. I investigate how this divergence affects the measured agreement of an LLM judge with human ratings: the same judge correlates \textbf{0.55 with experts and 0.04 with crowdworkers}. This gap is not readily explained by a weak judge: a three-expert panel predicts crowd ratings no better, and scaling the judge $64\times$ does not close the gap. A variance decomposition further shows that expert ratings distinguish between summaries produced by different systems, whereas crowd ratings vary mainly by source article. Similar divergence between expert and crowdsourced ratings has been documented in machine translation \citep{Freitag_2021}, where evaluation protocols also yield different system rankings. I extend this analysis to LLM judges, testing whether the gap persists across model scales and examining which sources of variation each panel captures.

\section{Method}
\looseness=-1

One judge scores every item, and those scores are correlated against each panel separately, so only
the label source varies. SummEval provides 1{,}600 summaries $=$ 16 systems $\times$ 100 CNN/DailyMail
articles, rated by three experts and five crowdworkers on coherence, consistency, fluency and
relevance (1--5), released as separate panels with disagreement unfiltered. Both panels received the
same instructions, but the experts rated in two rounds, the second with peers' first-round scores
visible, and the crowd in one anonymous round, so the panels differ in protocol as well as
personnel.

Judges are Qwen2.5-Instruct \citep{qwen2025qwen25technicalreport} from 0.5B to 32B, one family, so
capability is the only thing varying. Each rating is a single forward pass read from the logits over
the tokens \texttt{1}--\texttt{5} at the answer position as a probability-weighted expected value
\citep{liu2023gevalnlgevaluationusing}; there is no generation and no parsing
(Appendix~\ref{app:readout}). Reported are Pearson correlation against each panel's mean,
leave-one-rater-out baselines within each panel, and $\eta^2$, the variance share explained by a
grouping factor, against a permutation null; intervals bootstrap over items clustered by article.

\section{Results}

Values are for the 32B judge, the right-hand endpoint of Figure~\ref{fig:ladder}; per-dimension
results are in Appendix~\ref{app:results}. The judge correlates 0.433--0.605 with the expert panel
and 0.025--0.073 with the crowd panel, a difference of 0.405--0.581 with 95\% intervals excluding
zero on all four dimensions when clustered by source article. The crowd panel is nonetheless
measuring something: its ratings carry significant variance tied to the source article
($\eta^2 = 0.11$--$0.16$, permutation $p < 0.001$) and none tied to the system that wrote the
summary ($\eta^2 = 0.006$--$0.012$, $p \geq 0.17$, null expectation 0.009), where experts reach
0.18--0.34 at the system level. One crowdworker also predicts the rest of that panel at
0.613--0.628, so it is internally coherent.

The failure is not the judge's. A three-expert panel predicts the crowd labels as poorly as the
judge does (0.006--0.072 against 0.025--0.073), and across the ladder expert agreement rises
$0.133 \rightarrow 0.550$ while crowd agreement does not move (0.019--0.053). The judge inherits the expert
construct faithfully, and the crowd construct is orthogonal to it: the ceiling is a property of the
target, not a deficiency of the estimator.

\begin{figure}[t]
  \centering
  \includegraphics[width=0.92\linewidth]{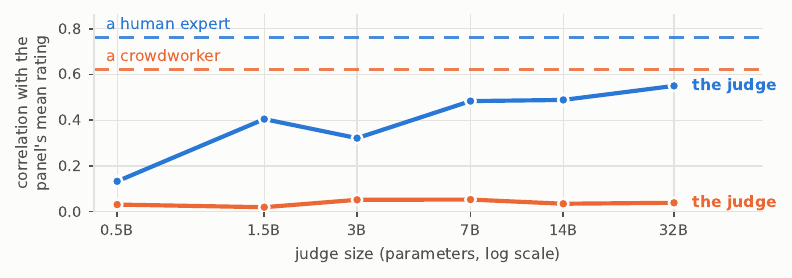}
  \caption{Judge--panel correlation ($y$) against judge size ($x$, log scale), averaged over the four
  dimensions. Colour is the panel: blue the 3 experts, orange the 5 crowdworkers. Solid lines are the
  judge; dashed lines are what one rater of that panel achieves against the rest of their own panel.}
  \label{fig:ladder}
\end{figure}

\section{Discussion and limitations}

Agreement with humans is a joint property of the judge and the panel: the same system reads as
strongly validated or as uncorrelated depending on who was asked. The field's safeguard does not
fire: both panel means are reliable ($R_3 = 0.79$, $R_5 = 0.83$), so any disattenuation declares
both targets sound. Reliability measures consistency among annotators, not whether their ratings capture the construct of interest. Correcting for annotator noise can improve estimates of that construct, but it cannot establish that the panel measures it in the first place. One
cheap diagnostic is $\eta^2$ against the comparison of interest, which fires where reliability does
not, though it is necessary rather than sufficient and presupposes that ranking systems is the
quantity of interest. The panels are shown to be non-interchangeable, not that either is right
(Appendix~\ref{app:impacts}).

The cause of the divergence is not identified here; the candidates are all properties of how the
annotation was commissioned, and separating them needs new collection. Only one dataset is tested,
and that is part of the finding: few evaluation resources release two identifiable annotator pools
over shared items at all, which is what this check requires (Appendix~\ref{app:audit}). The judge
also tracks summary length (Appendix~\ref{app:choices}), and one model family and one prompt
template are varied. 

\bibliographystyle{plainnat}
\bibliography{custom}

\newpage
\appendix
\section{Extended related work}
\label{app:related}

\paragraph{Precedent in machine translation.} the phenomenon investigated in this paper has a close precedent. \citet{Freitag_2021} re-annotated WMT test sets with professional translators using MQM and compared against the crowdsourced Direct Assesments scores the WMT test collects. They find the two protocols rank systems substantially differently, and that embedding-based automatic metrics can track the expert
ratings better than the crowd ratings do. 
That is my result --- an automatic scorer tracking
experts more closely than crowdworkers --- in another domain, five years earlier. What I add is the
LLM-judge setting, the separation of reliability from validity, the system-level signal measurement,
and the demonstration that judge capability does not close the gap.

\paragraph{The divergence is the dataset's result}
\citet{fabbri2021summevalreevaluatingsummarizationevaluation} compare the two panels directly: they
averaged scores within each annotator group and computed Pearson correlation, and report that it
``returned a value close to 0, indicating no correlation between expert and crowd-sourced judges.''
They then use the expert annotations throughout ``to avoid possible quality problems associated with
crowd-sourced ratings.'' They also report Krippendorff's $\alpha$ of 0.4920 for crowdworkers against
0.4132 for expert round one and 0.7127 for round two, and document a halo effect: crowd annotations
``tend to be similar across quality dimensions even when distinctions exist.'' This is my premise,
and I state it as such. What I add is what the divergence does to a judge.

\paragraph{Judge validation and label structure.}
\citet{guerdan2025validatingllmasajudgesystemsrating} give an identifiability result for judge
validation on the axis of elicitation format, forced choice versus response set; my axis is
different --- which population produced the label, not how it was elicited.
\citet{elangovan2025correlationimpacthumanuncertainty} stratify judge--human correlation by human
certainty and show the comparison flatters the judge on uncertain items; they use SummEval's expert
annotations only, and their reported $\alpha$ values are what I validate my data copy against.
\citet{liu2023gevalnlgevaluationusing} is my rubric lineage, and
\citet{bavaresco2025llmsinsteadhumanjudges} estimate human ceilings for judges by bootstrap.

\paragraph{Disagreement and reliability.} \citet{plank2022problemhumanlabelvariation} and
\citet{nie2020learncollectivehumanopinions} argue annotator disagreement is signal rather than noise.
The identity ceiling $=\sqrt{R_k}$ is due to \citet{cumlin2026rhoperfectcorrelationceilingsubjective}
and is cited rather than claimed; I note it is a single-author preprint. Classical reliability
follows Lord and Novick (1968) and $\alpha$ follows Krippendorff (2011).

\section{Data provenance and verification}
\label{app:provenance}

SummEval's canonical distribution point now requires authentication and returns HTTP 403 as of
30 August 2026. I use the already-paired release mirrored in the repository accompanying
\citet{liu2024aligning}. Because this is not the canonical host, I verify it.

\paragraph{Structural checks.} 1{,}600 rows $=$ 16 systems $\times$ 100 CNN/DailyMail articles;
exactly three expert and five crowd annotations per item; all ratings integers in 1--5; the source
article present in each record (median 2{,}180 characters) with 11 references.

\paragraph{Statistical verification.} Krippendorff's ordinal $\alpha$ computed from my copy
reproduces two independently published sets of values. My crowd values (coherence 0.499,
consistency 0.506, fluency 0.492, relevance 0.485) average 0.4955 against SummEval's published
0.4920. My expert coherence (0.554) and relevance (0.397) match the 0.55 and 0.40 reported by
\citet{elangovan2025correlationimpacthumanuncertainty}. Because the crowd values do not match, that
work evidently used the expert annotations, which its caption does not state.

\paragraph{The mirror adds source-article text that the canonical release withholds.} SummEval's own
documentation states the data is shared without the source articles used to generate the summaries.
The mirrored copy populates a text field on all 1{,}600 rows with the full article. I consume this
text to build judge prompts; I do not redistribute it. The mirror's MIT licence covers its own
material and cannot convey rights in the underlying journalism.

\paragraph{Two warnings for reproduction.} The \texttt{mteb/summeval} release ships one averaged
score per system per article, destroying the annotator dimension; this analysis cannot be run on it.
The \texttt{KnutJaegersberg/summeval\_pairs} release does preserve the per-annotator lists but
carries 1{,}700 rows across 17 systems rather than 1{,}600 across 16. Separately, the \texttt{id}
field is the source-article identifier and repeats across all 16 systems; keying items on it alone
silently collapses 1{,}600 items to 100.

\section{Judge prompting and readout}
\label{app:readout}

\paragraph{Prompt.} The system message instructs the model to rate summaries on a 1--5 scale and
reply with a single digit. The user message states the task, gives the dimension's criteria, the
source article and the summary, and closes by asking for a single digit. Criteria are
G-Eval-style re-implementations \citep{liu2023gevalnlgevaluationusing}, not verbatim. Articles are
truncated at 6{,}000 characters; typical prompts are around 1{,}800 tokens.

\paragraph{Readout.} I do not generate. A single forward pass is run over the chat-templated
prompt, and the rating is read from the logits over the tokens \texttt{1}--\texttt{5} at the answer
position. I report the probability-weighted expected value $\sum_r r \cdot P(r)$ over those five
tokens after renormalisation, recording the argmax alongside. This removes decoding noise, parsing
failure, and any possibility of an off-by-one in locating the rating token.

\paragraph{Validity of the readout.} I record the total probability the model places on the five
digit tokens before renormalisation. Across all runs the minimum observed is 0.9997 and the median
is 1.0000, so the renormalised distribution is not an artefact of the model intending some other
continuation. I verify separately that \texttt{1}--\texttt{5} are single tokens for this tokenizer;
the pipeline fails loudly if not. Padding is left-side so that the final position is the true last
token of every sequence, and truncation is also left-side so that an over-long prompt loses the
opening of the article rather than the rubric and the answer position.

\section{Reliability calculation}
\label{app:reliability}

Single-rater reliability $R_1$ is estimated by parallel forms, the correlation between two disjoint
single-rater draws, and $R_k$ by Spearman--Brown. The achievable-correlation ceiling $\sqrt{R_k}$
follows \citet{cumlin2026rhoperfectcorrelationceilingsubjective}.

\begin{table}[h]
  \caption{Panel reliability. The crowd panel has five raters against the experts' three, so a raw
  ceiling comparison confounds panel identity with panel size; the final column holds both at
  $k=3$, where experts lead on three of four dimensions. Both panels' aggregates are reliable by any
  conventional standard, which is the point of Section~4.}
  \label{tab:reliability}
  \centering
  \small
  \begin{tabular}{llccccc}
    \toprule
    Dimension & Pool & $k$ & $R_1$ & $R_k$ & ceiling & ceiling at $k=3$ \\
    \midrule
    Coherence & expert & 3 & 0.559 & 0.792 & 0.890 & 0.890 \\
    Coherence & crowd & 5 & 0.493 & 0.829 & 0.911 & 0.863 \\
    Consistency & expert & 3 & 0.900 & 0.964 & 0.982 & 0.982 \\
    Consistency & crowd & 5 & 0.496 & 0.831 & 0.912 & 0.864 \\
    Fluency & expert & 3 & 0.727 & 0.889 & 0.943 & 0.943 \\
    Fluency & crowd & 5 & 0.484 & 0.824 & 0.908 & 0.859 \\
    Relevance & expert & 3 & 0.453 & 0.713 & 0.845 & 0.845 \\
    Relevance & crowd & 5 & 0.480 & 0.822 & 0.907 & 0.857 \\
    \bottomrule
  \end{tabular}
\end{table}

\section{Full results behind Figure~\ref{fig:ladder}
\label{app:results}}

\begin{table}[t]
  \caption{Pearson correlation between each pair of raters, per SummEval dimension, over all 1{,}600 summaries. The judge is Qwen2.5-32B-Instruct, the largest in Figure~\ref{fig:ladder}. Columns 1--2 score the judge against each human panel's mean rating: first the 3-expert panel, then the 5-crowdworker panel. Columns 3--4 replace the judge with a human: column 3 is the expert panel's mean against the crowd panel's mean; column 4 is one crowdworker against the mean of the remaining four (leave-one-out). Judge--expert agreement runs 0.433--0.605 and judge--crowd 0.025--0.073, while expert--crowd runs 0.006--0.072 and crowd--crowd 0.613--0.628.}
  \label{tab:controls}
  \centering
  \small
  \begin{tabular}{lcccc}
    \toprule
    & \multicolumn{2}{c}{judge scored against} & \multicolumn{2}{c}{human scored against} \\
    \cmidrule(lr){2-3} \cmidrule(lr){4-5}
    Dimension & experts & crowd & crowd (by experts) & crowd (by crowd) \\
    \midrule
    Coherence & 0.568 & 0.030 & 0.040 & 0.627 \\
    Consistency & 0.605 & 0.025 & 0.006 & 0.628 \\
    Fluency & 0.433 & 0.028 & 0.052 & 0.618 \\
    Relevance & 0.594 & 0.073 & 0.072 & 0.613 \\
    \bottomrule
  \end{tabular}
\end{table}

\begin{table}[t]
  \caption{Full per-dimension values behind Figure~\ref{fig:ladder}, which plots the Mean column of the first two blocks. Rows are the six Qwen2.5-Instruct judges; columns are SummEval's four dimensions (coherence, consistency, fluency, relevance) and their mean, over all 1{,}600 summaries. The first block is Pearson $r$ between the judge and the 3-expert panel's mean rating; the second is the same judge against the 5-crowdworker panel's mean. The third is $\eta^2$, the share of score variance explained by which of the 16 summarisation systems produced the summary. Italic rows give the human references: one rater against the mean of the rest of their own panel (leave-one-out) in the first two blocks, and each panel's own $\eta^2$ in the third. Judge--expert agreement rises from 0.133 to 0.550 across the ladder while judge--crowd agreement stays between 0.019 and 0.053; judge $\eta^2$ rises from 0.055 to 0.395 against 0.006--0.012 for the crowd panel.}
  \label{tab:a1}
  \centering
  \small
  \begin{tabular}{lccccc}
    \toprule
    & Coh. & Cons. & Flu. & Rel. & Mean \\
    \midrule
    \multicolumn{6}{l}{\textit{Judge vs.\ expert panel ($r$)}} \\
    \quad 0.5B & 0.099 & 0.057 & 0.087 & 0.288 & 0.133 \\
    \quad 1.5B & 0.378 & 0.425 & 0.355 & 0.461 & 0.405 \\
    \quad 3B & 0.229 & 0.433 & 0.206 & 0.418 & 0.321 \\
    \quad 7B & 0.373 & 0.637 & 0.393 & 0.531 & 0.484 \\
    \quad 14B & 0.476 & 0.641 & 0.265 & 0.572 & 0.489 \\
    \quad 32B & 0.568 & 0.605 & 0.433 & 0.594 & 0.550 \\
    \quad \textit{one expert, vs.\ rest of panel} & 0.722 & 0.934 & 0.794 & 0.595 & 0.761 \\
    \midrule
    \multicolumn{6}{l}{\textit{Judge vs.\ crowd panel ($r$)}} \\
    \quad 0.5B & 0.018 & 0.051 & -0.007 & 0.063 & 0.031 \\
    \quad 1.5B & 0.036 & -0.012 & 0.029 & 0.025 & 0.019 \\
    \quad 3B & 0.051 & 0.053 & 0.041 & 0.062 & 0.052 \\
    \quad 7B & 0.046 & 0.049 & 0.047 & 0.069 & 0.053 \\
    \quad 14B & 0.019 & 0.039 & 0.011 & 0.069 & 0.035 \\
    \quad 32B & 0.030 & 0.025 & 0.028 & 0.073 & 0.039 \\
    \quad \textit{one crowdworker, vs.\ rest of panel} & 0.627 & 0.628 & 0.618 & 0.613 & 0.622 \\
    \midrule
    \multicolumn{6}{l}{\textit{System-level signal ($\eta^2$)}} \\
    \quad 0.5B & 0.054 & 0.051 & 0.050 & 0.064 & 0.055 \\
    \quad 1.5B & 0.215 & 0.239 & 0.235 & 0.223 & 0.228 \\
    \quad 3B & 0.114 & 0.318 & 0.210 & 0.289 & 0.233 \\
    \quad 7B & 0.294 & 0.367 & 0.317 & 0.281 & 0.315 \\
    \quad 14B & 0.263 & 0.384 & 0.286 & 0.282 & 0.304 \\
    \quad 32B & 0.381 & 0.449 & 0.425 & 0.323 & 0.395 \\
    \quad \textit{expert panel mean} & 0.300 & 0.339 & 0.242 & 0.179 & 0.265 \\
    \quad \textit{crowd panel mean} & 0.006 & 0.010 & 0.012 & 0.012 & 0.010 \\
    \bottomrule
  \end{tabular}
\end{table}

Three observations the dimension means in Figure~\ref{fig:ladder} conceal. Judge--crowd agreement is
flat at the cell level, not merely on average: all 24 cells fall between $-0.007$ and 0.073, and no
dimension or model size breaks out. Relevance sits at the top of the crowd range at every rung
(0.063, 0.025, 0.062, 0.069, 0.069, 0.073), but I do not read this as signal: crowd relevance
$\eta^2$ is 0.012 against a permutation null median of 0.009 ($p = 0.183$). And the ladder is
non-monotone at 3B on three of four dimensions: coherence $0.378 \rightarrow 0.229$, fluency
$0.355 \rightarrow 0.206$, relevance $0.461 \rightarrow 0.418$, and at 14B on fluency; the 3B
judge is worse than the 1.5B judge on the dimension mean. I attribute this to prompt sensitivity
and report it rather than smoothing it.

Judge $\eta^2$ rises $0.055 \rightarrow 0.395$ across the ladder, overtaking the expert panel's mean
of 0.265 between 7B and 14B, while the crowd panel's own $\eta^2$ is 0.010. For the 32B judge, the
expert-minus-crowd correlation difference, bootstrapped 2{,}000 times over items clustered by source
article, is: coherence 0.538 $[+0.471, +0.599]$; consistency 0.581 $[+0.517, +0.649]$; fluency 0.405
$[+0.337, +0.471]$; relevance 0.521 $[+0.461, +0.583]$.

\section{Controls used}
\label{app:controls}

Each rules out an alternative reading of the judge--crowd gap.

\paragraph{Not annotator noise.} Crowdworkers predict one another at 0.613--0.628 (leave-one-rater-out
against the mean of the remaining four). On relevance the crowd panel agrees with itself more than
the expert panel does (0.613 versus 0.595).

\paragraph{Not a halo effect.} SummEval attributes crowd behaviour to overall quality being smeared across dimensions. If that were the explanation, crowd ratings should track the experts' overall
score even where they miss the specific dimension. But they do not: crowd-overall against expert-overall is 0.068, and per dimension the gain from comparing against expert-overall rather than the matched dimension is between $-0.023$ and $+0.044$. The halo is mildly real (Mean inter-dimension
correlation within the crowd panel is 0.543 against 0.438 for experts) and nowhere near
sufficient.

\paragraph{Not summary length.} Crowd ratings correlate with summary length at 0.032, 0.054,
$-0.006$ and 0.087 across the four dimensions. Summary length does separate systems strongly
($\eta^2$ of log length by system $= 0.339$) and my judge does track length, so judge $\eta^2$
could in principle be a length effect; partialling log length out leaves judge $\eta^2$ at 0.369
(from 0.395) and expert $\eta^2$ at 0.261 (from 0.265).

\paragraph{Not judge weakness.} Expert panel minus judge, at predicting the crowd aggregate:
$+0.006$, $+0.043$, $-0.004$, $-0.003$.

\paragraph{Not the readout, and not sample size.} The headline pattern replicates using the argmax
rating in place of the probability-weighted expected value, and a 100-item pilot reproduced the
direction and rough magnitude before the full 1{,}600-item run.

\section{Dataset audit}
\label{app:audit}

My experiment requires two or more identifiable annotator pools rating the same items, with
per-pool labels released uncollapsed. I audited 13 judge-validation resources against eight
structural criteria, checking the current release rather than the describing paper: (1) at least two
identifiable pools; (2) pools rate shared items; (3) per-pool labels uncollapsed; (4) at least 300
items; (5) at least three raters per pool per item; (6) source content released; (7) ordinal or
continuous ratings; (8) a system or condition variable enabling $\eta^2$.

\begin{table}[h]
  \caption{Structural audit of judge-validation resources. SummEval is the only resource that
  supports this check at item level; WMT supports it at system level only. SummEval's qualified
  mark on (6) is the canonical/mirror split described in Appendix~B.}
  \label{tab:audit}
  \centering
  \small
  \begin{tabular}{lcccccccc l}
    \toprule
    Dataset & 1 & 2 & 3 & 4 & 5 & 6 & 7 & 8 & Detects pool disagreement? \\
    \midrule
    SummEval & \checkmark & \checkmark & \checkmark & \checkmark & \checkmark & $\sim$ & \checkmark & \checkmark & \textbf{yes, item level} \\
    WMT MQM/DA & \checkmark & \checkmark & $\sim$ & \checkmark & --- & \checkmark & \checkmark & \checkmark & system level only \\
    DICES-350 & \checkmark & \checkmark & \checkmark & \checkmark & \checkmark & \checkmark & $\sim$ & --- & no, fails 8 \\
    DICES-990 & \checkmark & \checkmark & \checkmark & \checkmark & \checkmark & \checkmark & $\sim$ & --- & no, fails 8 \\
    HANNA & --- & --- & \checkmark & \checkmark & $\sim$ & \checkmark & \checkmark & \checkmark & no, fails 1 \\
    ChaosNLI & --- & --- & --- & \checkmark & \checkmark & \checkmark & $\sim$ & --- & no, fails 1 and 3 \\
    MT-Bench & $\sim$ & $\sim$ & \checkmark & --- & $\sim$ & \checkmark & --- & \checkmark & no, fails 4 and 7 \\
    Chatbot Arena & --- & --- & --- & \checkmark & --- & \checkmark & --- & \checkmark & no \\
    HelpSteer2 & --- & --- & $\sim$ & \checkmark & $\sim$ & \checkmark & \checkmark & --- & no, fails 1 \\
    HelpSteer3 & --- & --- & \checkmark & \checkmark & $\sim$ & \checkmark & \checkmark & --- & no, fails 1 \\
    JudgeBench & --- & --- & --- & \checkmark & --- & \checkmark & --- & \checkmark & no, by design \\
    RewardBench & --- & --- & --- & \checkmark & --- & \checkmark & --- & \checkmark & no \\
    PandaLM & --- & --- & \checkmark & \checkmark & \checkmark & \checkmark & $\sim$ & \checkmark & no, fails 1 \\
    \bottomrule
  \end{tabular}
\end{table}

Criterion (2) eliminates most candidates and (8), which enabled the analysis that mattered most,
eliminates several of the rest. WMT's MQM annotations carry per-segment rater identity, but Direct
Assessment collects roughly one to two ratings per segment because it is built for system-level
aggregation. DICES has excellent pool structure ---
123 raters, a complete design, every pool rating every item --- and fails only because all responses
come from a single generator. HANNA \citep{chhun2022humancriteriaautomaticmetrics} satisfies (8)
with 11 generators and fails on (1) alone. ChaosNLI \citep{nie2020learncollectivehumanopinions}
releases label counts without annotator identifiers, so pools cannot be formed at all. HelpSteer2 and HelpSteer3
release the three most-agreeing of up to five annotators, which makes $R_k$ run optimistic without
discarding items. I excluded ASAP-AES: its distribution sits behind competition terms I could not
audit, and it fails (1), (5) and (8) on any published account.

This scarcity is not incidental to the result. A benchmark that merges its annotators into one gold
column has destroyed the evidence needed to notice that the merge mattered.

\section{Alternate datasets}
\label{app:choices}

\paragraph{Why not DICES.} It has genuine pool structure but every response comes from a single
model, so there is no system variable and $\eta^2$ cannot be computed. My own prior analysis found
that demographic strata there differ in where the safety threshold sits rather than in which items
they flag, which predicts a null for the pool contrast.

\paragraph{Why not WMT.} MQM is released at segment level with professional raters, but Direct
Assessment lacks the per-segment rater depth, so the item-level analysis cannot be built. The
system-level comparison is possible and is where \citet{Freitag_2021} already reported the
divergence.

\paragraph{Why not HANNA.} Twenty-two raters at three per item with no pool structure and a badly
unbalanced design.

\section{Compute and reproduction}
\label{app:compute}

All judge inference ran on serverless GPUs. Rungs up to 7B ran on a single A10G at batch size 16;
the 14B rung on an A100-40GB; the 32B rung on an A100-80GB at batch size 8. Total inference across
all six rungs and 6{,}400 prompts per rung was under two GPU-hours. All statistical analysis runs on
CPU in minutes.



\paragraph{Prompt sensitivity and annotator-facing rubrics.} Three rubric paraphrases per dimension
would give a sensitivity band per rung and would test whether the non-monotonicity above is
template-driven. Separately, I prompt the judge with G-Eval-style criteria, but the comparison is
against SummEval's annotators; prompting the judge with the instructions those annotators actually
saw would be the stricter like-for-like test.

\section{Broader impacts}
\label{app:impacts}

I report that an LLM judge's agreement with humans is a joint property of the judge and the
annotator panel: the same scores on the same 1{,}600 SummEval summaries correlate 0.55 with the
dataset's three-expert panel and 0.04 with its five-crowdworker panel. The intended benefit is cheap
detection of invalid evaluation targets before they gate model releases. Reporting which panel a
judge was validated against, and whether that panel discriminates the comparison the benchmark
exists to make, costs nothing beyond data already collected and fires here where internal-reliability
checks do not.

The misuse I am most concerned about is reading this as evidence that crowdworker annotation is
low quality. It is not. SummEval's crowd annotators were screened, restricted to native English
speakers and paid an average of USD 12 per hour; their aggregate is reliable ($R_5 = 0.83$); and
their ratings carry significant variance tied to the source article ($\eta^2 = 0.11$--$0.16$,
permutation $p < 0.001$) while carrying none tied to the system that produced the summary
($\eta^2 = 0.006$--$0.012$, $p \geq 0.17$). They measured something, consistently. It was not the
contrast the benchmark ranks. The two panels also differ in protocol and not only in personnel ---
the expert ratings went through a second, peer-visible adjudication round that the crowd ratings did
not --- so this result speaks to how evaluation labels are commissioned, licenses no inference about
annotator skill, and recommends no change to who is hired.

\section{Variance decomposition}
\label{app:variance}

Judge is Qwen2.5-32B-Instruct, probability-weighted readout, all 1{,}600 items; permutation nulls
are 2{,}000 iterations. The null expectation of $\eta^2$ is 0.009 for the 16-system factor and 0.062
for the 100-article factor.

\begin{table}[h]
  \caption{Where each rater's signal lives. Every crowd cell is at the permutation null at the
  system level and far above it at the article level: crowd raters respond to properties shared by
  all sixteen summaries of an article rather than to differences between them. Expert consistency is
  the only human cell with no article-level signal, which is what the construct definition requires.}
  \label{tab:variance}
  \centering
  \small
  \begin{tabular}{llcccc}
    \toprule
    Rater & Dimension & $\eta^2$ by system & $p$ & $\eta^2$ by article & $p$ \\
    \midrule
    judge & coherence & 0.381 & $<$0.001 & 0.161 & $<$0.001 \\
    judge & consistency & 0.449 & $<$0.001 & 0.128 & $<$0.001 \\
    judge & fluency & 0.425 & $<$0.001 & 0.140 & $<$0.001 \\
    judge & relevance & 0.323 & $<$0.001 & 0.178 & $<$0.001 \\
    \midrule
    expert & coherence & 0.300 & $<$0.001 & 0.144 & $<$0.001 \\
    expert & consistency & 0.339 & $<$0.001 & 0.054 & 0.818 \\
    expert & fluency & 0.242 & $<$0.001 & 0.089 & 0.001 \\
    expert & relevance & 0.179 & $<$0.001 & 0.232 & $<$0.001 \\
    \midrule
    crowd & coherence & 0.006 & 0.812 & 0.128 & $<$0.001 \\
    crowd & consistency & 0.010 & 0.409 & 0.155 & $<$0.001 \\
    crowd & fluency & 0.012 & 0.172 & 0.109 & $<$0.001 \\
    crowd & relevance & 0.012 & 0.183 & 0.152 & $<$0.001 \\
    \bottomrule
  \end{tabular}
\end{table}

\section{Asset licences}
\label{app:licences}

SummEval's annotations are MIT-licensed. The canonical host now returns HTTP 403, so this
paper uses the copy redistributed in the PairS repository \citep{liu2024aligning},%
\footnote{\url{https://github.com/cambridgeltl/PairS}, file~\path{data/SummEval/model_annotations.aligned.paired.jsonl}.}
which carries its own MIT licence over its own material. That copy also supplies the
CNN/DailyMail article text the canonical SummEval release withholds; the Apache-2.0 terms
attached to CNN/DailyMail cover the dataset packaging rather than the underlying
journalism, and the text is consumed here to build prompts rather than redistributed.
Qwen2.5-Instruct is Apache-2.0 at every scale used except 3B, which is research-only under
\texttt{qwen-research}. Audit-only datasets are not used, modified or redistributed by
this work.

\newpage

\end{document}